\documentclass[preprint,12pt]{elsarticle}

\usepackage{amssymb}
\usepackage{amsmath}
\usepackage{booktabs}
\usepackage{multirow}
\usepackage{graphicx}  

\journal{Engineering Analysis with Boundary Elements}

\begin{document}

\begin{frontmatter}



\title{Deep Parallel Spectral Neural Operators for Solving Partial Differential Equations with Enhanced Low-Frequency Learning Capability}


\author[a,b,c]{Qinglong Ma} 
\author[a,b,c]{Peizhi Zhao}
\author[a,b,c]{Sen Wang} 
\author[a,b,c]{Tao Song\corref{*}} 

\affiliation[a]{organization={College of Computer Science and Technology},
            addressline={China University of Petroleum (East China)}, 
            city={Qingdao},
            postcode={266580}, 
            state={Shandong},
            country={China}}
\affiliation[b]{organization={State Key Laboratory of Chemical Safety},
}
\affiliation[c]{organization={Shandong Key Laboratory of Intelligent Oil \& Gas Industrial Software},
}

\cortext[*]{corresponding author}
\ead{tsong@upc.edu.cn}

\begin{abstract}
Designing universal artificial intelligence (AI) solver for partial differential equations (PDEs) is an open-ended problem and a significant challenge in science and engineering. Currently, data-driven solvers have achieved great success, such as neural operators. However, the ability of various neural operator solvers to learn low-frequency information still needs improvement. In this study, we propose a Deep Parallel Spectral Neural Operator (DPNO) to enhance the ability to learn low-frequency information. Our method enhances the neural operator's ability to learn low-frequency information through parallel modules. In addition, due to the presence of truncation coefficients, some high-frequency information is lost during the nonlinear learning process. We smooth this information through convolutional mappings, thereby reducing high-frequency errors. We selected several challenging partial differential equation datasets for experimentation, and DPNO performed exceptionally well. As a neural operator, DPNO also possesses the capability of resolution invariance.
\end{abstract}



\begin{keyword}


Mechanical partial differential equations, neural operator, scientific machine learning, deep learning-based PDE solver.
\end{keyword}

\end{frontmatter}



\section{Introduction}

In scientific and engineering fields, partial differential equations (PDEs) are widely used to model phenomena across various physical domains, such as fluid mechanics, materials science, quantum mechanics, and electromagnetism. Most PDEs, including the Navier-Stokes equation, lack analytical solutions, which positions numerical methods as the primary approach for solving them \citep{gockenbach2010partial,tanabe2017functional}. While these methods are effective, they are computationally intensive and often require substantial computational resources. Classical numerical approaches typically involve trade-offs between accuracy and computational efficiency. Consequently, there is an urgent need to identify a class of solution methods that balance both efficiency and accuracy.

In recent years, significant progress in deep learning has enabled artificial intelligence to potentially replace traditional numerical solvers and simulators \citep{raissi2019physics,esmaeilzadeh2020meshfreeflownet,greenfeld2019learning,kochkov2021machine}. Notably, the concept of neural operators and their innovative architectures has shown considerable promise in approximating complex mappings, with extensive research conducted in this area. This paradigm focuses on approximating the relationship between inputs and outputs by training deep models with novel architectures. Specifically, it captures correlations in time series data, such as predicting future flows and pressures based on historical flow and pressure data \citep{lim2021time}. Moreover, the method is capable of directly learning the mapping relationships between different parameters, such as inferring the macroscopic mechanical properties of a composite material from its microstructure \citep{sengodan2021prediction}. This approach has demonstrated exceptional performance across various fields, including fluid mechanics and material science.

The existing neural operators, while lacking the ability to learn low-frequency components effectively, also lack efficient handling of high-frequency information. As a result, the accuracy of neural operators remains insufficient in practical problems, limiting their application effectiveness. 

In practical problems, low-frequency components are often dominant over high-frequency components \citep{george2024incremental, boffetta2012two}. Thus, the primary error in neural operator learning arises from the insufficient ability to learn low-frequency components. Enhancing the neural operator's ability to capture low-frequency components is therefore key to building a more efficient solver. Furthermore, while low-frequency components capture large-scale trends, high-frequency information is crucial for modeling fine-grained local behaviors, especially in problems with complex boundary conditions and rapidly changing physical phenomena \citep{chamarthi2022importance}. 

In light of the challenges outlined above, we propose a solution to address these issues. Our contributions are as follows: We introduce a Deep Parallel Spectral Neural Operator (DPNO) that enhances the ability to learn low-frequency components. We design a parallel operator block that independently learns low-frequency components through dual branches, thereby improving its capability to learn low-frequency components. Furthermore, we introduce a projection network to smooth high-frequency information, enhancing the neural operator's ability to capture local details. In the experimental section, we select multiple challenging datasets, and the experiments demonstrate that DPNO performs well. We also conduct resolution invariance experiments.

\section{Related work}

\subsection{Neural operators}
\label{subsec1}

Neural operators are a series of methods for learning meshless, infinite-dimensional operators using neural networks that have been proposed in recent years of research. Neural operators solve the mesh dependence problem of finite-dimensional operator methods by generating a set of mesh parameters that can be used for different discretisations \citep{li2020neural}. It has the ability to transfer solutions between arbitrary meshes. In addition, the neural operator needs to be trained only once, and the subsequent performances of various new instance parameters require only forward propagation calculations. This reduces the computational cost associated with repeated training. Neural operators are purely data-driven and do not rely on any prior knowledge about partial differential equations \citep{kovachki2023neural}. This makes it possible for the operator to learn solutions to partial differential equations.

Specifically, neural operators are described as an iterative update architecture:  \( v_0 \rightarrow v_1 \rightarrow \cdots \rightarrow v_T \), Where \(v_j\) is a series of functions on \( \mathbb{R}^{d_v} \). Input a \( \in A \) is first elevated to a higher dimensional representation through local transformation \( P \): \( v_0(x) = P(a(x)) \). This local transformation is typically parameterized by a shallow fully connected neural network. Then apply several updates and iterations: \( v_t \rightarrow v_{t+1} \). The output \( u(x) = Q(v_T(x)) \) is the projection of \( v_T \) through local transformation \( Q \): \( \mathbb{R}^{d_v} \rightarrow \mathbb{R}^{d_u} \). Each iteration of \( v_t \rightarrow v_{t+1} \) is defined as a combination of non-local integration operator \( \mathcal{G} \) and local nonlinear activation function \( \sigma \). Among them, the local integration operator is formalized as:
\begin{equation}
    (G(a; \emptyset) v_t)(x) = \int_{D} \mathcal{K}(x, y, a(x), a(y); \emptyset) v_t(y) \, dy \quad \forall x \in D
\end{equation}
where \( D \) represents the domain of the function, \( \mathcal{K} \) represents the kernel integral operator, parameterized by \( \emptyset \).

Fourier transform is a mathematical transformation that analyzes the frequency components of a signal by decomposing it into a combination of sine and cosine waves of different frequencies. It has wide applications in many scientific and engineering fields \citep{kar2008vibration,sandryhaila2013discrete,candes2006robust}. In practice, it is a spectral method commonly used to solve partial differential equations \citep{canuto2007spectral,boyd2001chebyshev}.

In the Fourier domain, differential operations become simpler because differentiation is equivalent to multiplication in the Fourier domain. This makes the Fourier transform a powerful tool for solving partial differential equations, especially in cases involving frequency components. Through the Fourier transform, complex differential equations can be transformed into algebraic equations that are easier to handle, thereby simplifying the solving process. The Fourier transform has played an important role in the development of deep learning. In theory, the Fourier transform appears in the proof of the general approximation theorem, which states that multi-layer neural networks can approximate any continuous function \citep{hornik1989multilayer}. In practice, they are often used to accelerate convolutional neural networks \citep{mathieu2013fast}. Neural network architectures associated with Fourier transforms or sinusoidal activation functions have been proposed and studied \citep{bengio2007scaling,sitzmann2020implicit,mingo2004fourier}.

Recently, some spectral methods for partial differential equations have been extended to neural networks. A new neural operator architecture has been proposed by Li et al. \citep{li2020fourier}. This architecture directly defines a kernel integration operator in Fourier space with quasi-linear time complexity and advanced approximation capabilities. This advancement opens new avenues for the application of neural networks in meshless and infinite-dimensional spaces, thus enabling neural networks to efficiently handle complex partial differential equation problems.

\subsection{Spatial learning with attention mechanism}

Transformer \citep{waswani2017attention}, as an important cornerstone of deep learning, has recently been applied to solve partial differential equations. It directly learns the mapping relationships between operators in physical space and has achieved promising results. 

Liu et al. \citep{liu2022ht} combined the Swin Transformer \citep{liu2021swin} with the multigrid method to propose HT-Net, which captures multi-scale spatial correlations. Li et al. \citep{li2024scalable} proposed an efficient model, FactFormer, which utilizes a low-rank structure through multidimensional factorized attention. In addition, to address the quadratic complexity of attention, Li \citep{li2022transformer}, Hao \citep{hao2023gnot}, and Xiao et al. \citep{xiao2023improved} utilized existing linear Transformers to propose OFormer, GNOT, and ONO, significantly reducing the issue of computational inefficiency. However, the aforementioned methods still directly apply attention to a large number of mesh points. Therefore, Wu et al. \citep{wu2402transolver} proposed Transolver, which applies attention to the intrinsic physical states captured by learnable slices, thereby better modeling complex physical correlations. 

The integration of attention mechanisms into PDE solvers has demonstrated a marked improvement in handling complex spatial correlations. As these methods continue to evolve, they offer promising directions for more efficient and scalable solutions to high-dimensional problems.

\subsection{Spectral learning via spectral analysis techniques}

Many recent advancements in neural operators have focused on leveraging spectral transformations to solve partial differential equations efficiently. These spectral neural operators, which include a variety of methods that utilize Fourier transforms or similar techniques, have demonstrated strong performance in approximating and solving complex PDEs. For example, DeepONet, developed by Lu et al. \citep{lu2019deeponet}, is based on the universal approximation theorem. The network comprises two parts: a branch network and a trunk network, which process information from the sampled points of the input function and the target points, respectively. The outputs of these two parts are then combined to generate the final result, effectively approximating and predicting complex nonlinear operators.

For the irregular grid problem, Li et al. \citep{li2023fourier} proposed the Geo-FNO model. Geo-FNO works by transforming complex geometric tasks, such as point clouds, into a potentially unified mesh. This mesh is then used as an input to FNO to efficiently process complex geometries. Tran et al. \citep{tran2021factorized} proposed a decomposed Fourier neural operator (F-FNO). F-FNO introduces new separable spectral layers and improved residual connections, combined with training strategies such as the Markov assumption, Gaussian noise, and cosine learning rate decay. This significantly reduces errors on several benchmark PDEs. Gupta et al. \citep{gupta2021multiwavelet} proposed the Multi-Wavelet Transform Neural Operator (MWT). MWT leverages the fine-grained representation of the Multi-Wavelet Transform and orthogonal polynomial bases by compressing the kernel of the operator onto a Multi-Wavelet polynomial basis, demonstrating strong approximation capabilities in several PDE tasks. Zhang et al. \citep{fanaskov2023spectral} proposed a spectral neural operator (SNO), which converts inputs and outputs into coefficients of basis functions, learning the mapping between inputs and outputs sequentially while alleviating systematic bias caused by aliasing errors. Xiong et al. \citep{xiong2024koopman} proposed the Koopman Neural Operator (KNO) based on Koopman theory. The KNO replaces the feature mapping in the Fourier Neural Operator with the Koopman Operator, which improves the performance of FNO in long-term predictions.

In addition, there are other operator methods with deep architecture designs. Based on the U-Net concept and residual connections, multi-scale processing is performed for PDE tasks \citep{ronneberger2015u,szegedy2017inception}. For example, Zhang et al. \citep{rahman2022u} designed a U-shaped neural operator (U-NO). The U-shaped structure of U-NO improves model depth and memory efficiency. In benchmark tests for solving partial differential equations, such as Darcy's and Navier-Stokes equations, U-NO significantly improves prediction accuracy. Wen et al. \citep{wen2022u} integrated FNO with U-Net to design an enhanced Fourier neural operator (U-FNO). U-FNO enhances the ability of FNO to handle multi-scale data by using residual connections. Wu et al. \citep{wu2023solving} proposed the latent spectrum model (LSM). LSM introduces a projection technique that sequentially projects high-dimensional data into multiple compact latent spaces, thereby eliminating redundant coordinate information. Based on triangular basis operators, LSM decomposes complex nonlinear operators into multiple basis operators through neural spectral blocks, thereby enhancing the ability to solve complex partial differential equations. In most Fourier neural operators and their variants, the presence of fixed Fourier modes limits their ability to learn low-frequency components, whereas DPNO successfully alleviates this issue.

\section{Proposed method}

In this section, we primarily present the setup of the PDE problem, the architecture of DPNO, and its associated methods, which are composed of three core components and several standard neural operator components.

\subsection{Problem setup}
For a PDE-governed task, consider a bounded open set \( D \subset \mathbb{R}^d \) representing the coordinate domain. In this setting, the inputs and outputs can be represented as functions defined over the coordinate domain \( D \), which belong to the Banach spaces \( X \) and \( Y \), where \( X = X(D; \mathbb{R}^{dx}) \) and \( Y = Y(D; \mathbb{R}^{dy}) \). Here, \( \mathbb{R}^{dx} \) and \( \mathbb{R}^{dy} \) denote the ranges of the input and output functions, respectively.

At each coordinate \( s \in D \), the function values \( x(s) \in \mathbb{R}^{dx} \) and \( y(s) \in \mathbb{R}^{dy} \) represent the input and output values at location \( s \), analogous to pixel values. The PDE-solving process is thus described as approximating the optimal mapping operator \( G: X \to Y \), typically using a deep neural network model \( G_\theta \) to approximate this mapping, where \( \theta \in \Theta \) represents the parameter set of the model. The model learns from observed samples \( \{(x, y)\} \) to approximate \( G \).

\subsection{Fourier neural operator}
The convolution operator is introduced in FNO to replace the kernel integration operator in Equation 1. Let \( F \) denote the Fourier transform of a function \( f : D \to \mathbb{R}^d \), and \( F^{-1} \) its inverse. Then,
\begin{equation}
    (\mathcal{F}f)_j(k) = \int_{-D}^{D} f_j(x) e^{-2\pi i x k} \, dx
\end{equation}
And the inverse Fourier transform is given by:
\begin{equation}
    (\mathcal{F}^{-1}f)_j(x) = \int_{-D}^{D} f_j(k) e^{2\pi i x k} \, dk
\end{equation}
By specifying \( k(x, y, a(x), a(y)) = k(x, y) \), and applying the convolution theorem, the Fourier kernel integral operator can be obtained:
\begin{equation}
    K(\phi) v_t(x) = \mathcal{F}^{-1} \left( \mathcal{R}_\phi \cdot \left( \mathcal{F} v_t \right)(x) \right) \quad \forall x \in D
\end{equation}

In Fourier neural operators, mode truncation is used to limit the number of Fourier modes in the function's output. This is achieved by directly parameterizing the kernel function in the Fourier space, allowing only a finite number of low-frequency modes to be considered during convolution computation.

\subsection{DPNO architecture}

\begin{figure}[t] 
\centering 
\resizebox{\textwidth}{!}{
  \includegraphics{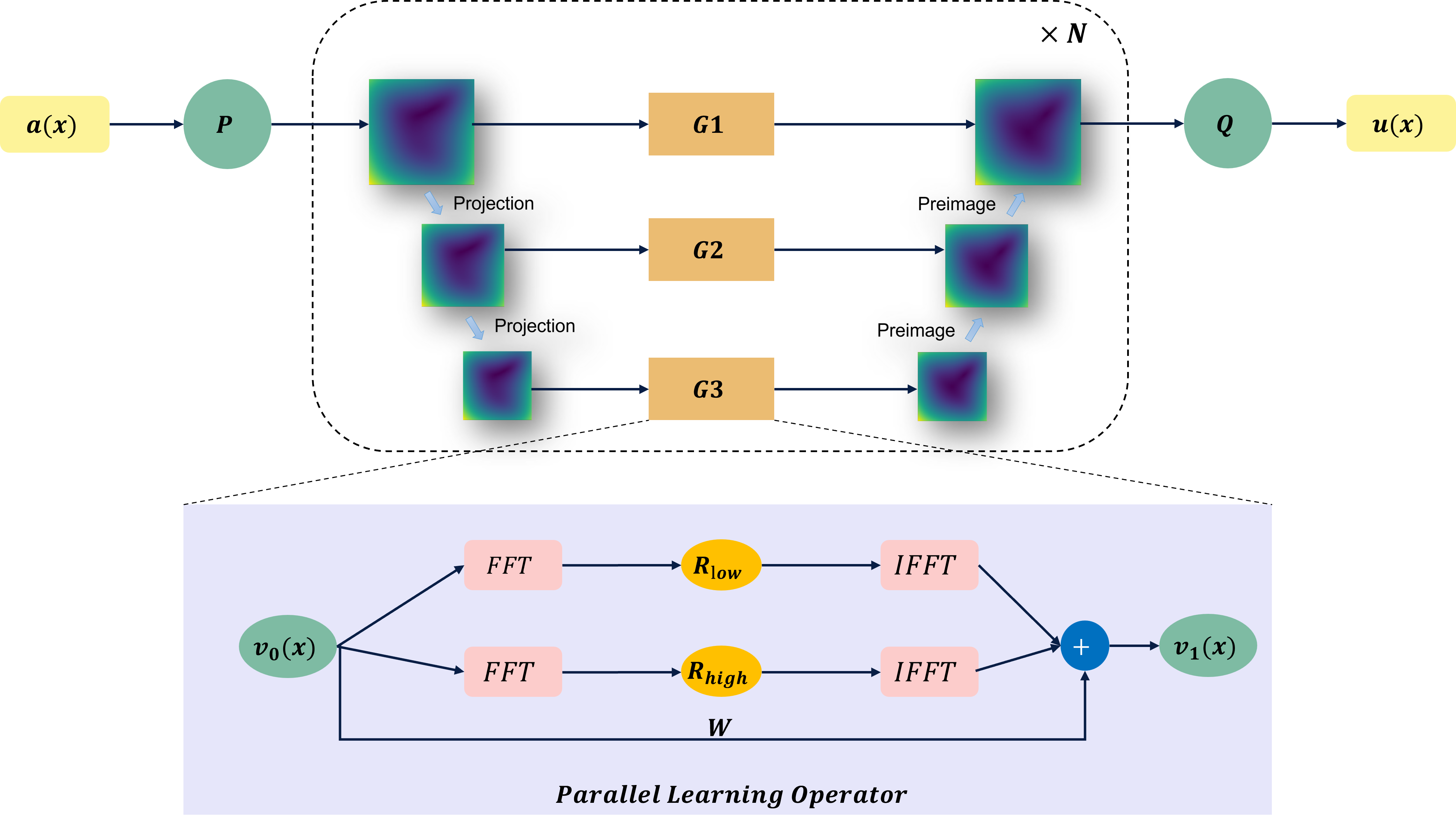} 
}
\caption{DPNO overall architecture}
\label{fig1} 
\end{figure}

DPNO belongs to a multi-scale neural operator architecture, as shown in Fig. \ref{fig1}. Specifically, it can be divided into the following four parts:

\begin{enumerate}
    \item The data preprocessing in this architecture follows the standard components used in previous neural operators. Given an input function \( a: \mathcal{D}_A \rightarrow \mathbb{R}^{d_A} \), a dimensionality-increasing operation \( P \) is first performed on \( a \) to compute \( v_0: \mathcal{D}_A \rightarrow \mathbb{R}^{d_0} \), where the operation \( P \) is parameterized by a shallow, fully connected neural network. Its function is as follows:
    \begin{equation}
        v_0(x) = P(a(x)) \quad \forall x \in \mathcal{D}_0
    \end{equation}
    where \( d_0 \gg d_A \).

    \item Before performing frequency domain learning, we use a standard encoder-decoder structure, which is composed of standard convolution and pooling operations. In the encoder part, downsampling is performed to suppress high-frequency information. The decoder part is the opposite of the encoder, where upsampling is used to restore the feature dimensions. The encoder and decoder are defined as:
    \begin{equation}
    v_i^{(L_i)} = \phi_i(v_i(x)) \quad i = 0, 1, \ldots, T-1
    \end{equation}
    where \( v_i^{(L_i)} \) represents the downsampled version of \( v_i \) in the \(i\)-th level, and \( \phi_i \) is the downsampling operation at that level.
    
    \begin{equation}
    v_i^{(U_i)} = \psi_i(v_i(x)) \quad i = 0, 1, \ldots, T-1
    \end{equation}
    where \( v_i^{(U_i)} \) represents the upsampled version of \( v_i \) in the \(i\)-th level, and \( \psi_i \) is the upsampling operation at that level.

    \item The parallel operator layer is a core component for enhancing the operator's ability to learn low-frequency components. It processes low-frequency information in different modes through two sub-layers and integrates the features by linearly combining the outputs of both sub-layers. Specifically, the data representations from different latent spaces, obtained through the encoder, undergo parallel Fourier transforms to capture frequency information in different modes. Additionally, a linear transformation is applied for residual connections to further supplement the high-frequency information. The formula is as follows:
    \begin{equation}
    G_{\theta_K} = \sum \left( R K_1, R K_2, W_{\theta} \right)
    \end{equation}
    Where RK is equivalent to the Fourier kernel integral operator in Equation (4). W represents a linear transformation.

    \item The projection operator \( Q \): The features processed by the decoder are mapped to the original physical space through the projection \( Q \). \\
    This is defined as:
    \begin{equation}
        u(x) = Q(v_1(x))
    \end{equation}

    where \( Q \) is a fully connected neural network.

\end{enumerate}

\section{Experiments}

\subsection{Experiment setup}

In this study, all experiments were performed on a Nvidia GeForce RTX 3090 24GB GPU with Python 3.9 and CUDA 12.3. To ensure the convergence of all models, we set the number of training rounds to 500. Each model parameter is updated using the Adam optimizer with a learning rate of 0.001 and weight decay of 0.0001. The L2 loss function (Mean Squared Error, MSE) is used as the optimization objective, which is consistent with all the baselines.

\subsection{Benchmarks}
In this section, we primarily introduce the dataset and baseline models used in this study.

\subsubsection{Baseline}
In order to better demonstrate the performance of our proposed model, we compare DPNO with eight currently recognized advanced models, including U-Net \citep{ronneberger2015u,szegedy2017inception}, FNO \citep{li2020fourier}, WMT \citep{gupta2021multiwavelet}, U-FNO \citep{wen2022u}, UNO \citep{rahman2022u}, F-FNO \citep{tran2021factorized}, SNO \citep{fanaskov2023spectral} and LSM \citep{wu2023solving}. This includes FNO and its variants, as well as other novel spectral methods.

\subsubsection{Dataset}
To comprehensively evaluate the performance of the model, we selected six benchmark datasets for partial differential equations and recorded the relative errors of the model. These include two solid mechanics-related partial differential equations and four fluid mechanics-related partial differential equations, as detailed in Table \ref{tab1}. The following presents the specific forms of six equations and their format translation.

\begin{table}[t]
\centering
\caption{Overview of Experiment Datasets and Benchmark Details.}
\resizebox{\textwidth}{!}{
\begin{tabular}{l l l l l l l} 
\toprule
Physics & Benchmark & Geometry & \#DIM & Task & Train Size & Test Size \\ 
\midrule
SOLID & Elasticity-G & Regular Grid & 2D & ESTIMATE STRESS & 1000 & 200\\
SOLID & Plasticity & Structured Mesh & 3D & MODEL DEFORMATION & 900 & 80\\
FLUID & Navier-Stokes & Regular Grid & 2D & PREDICT FUTURE & 1000 & 200\\
FLUID & Darcy & Regular Grid & 2D & ESTIMATE PRESSURE & 1000 & 200\\
FLUID & Airfoil & Structured Mesh & 2D & ESTIMATE VELOCITY & 1000 & 200\\
FLUID & Pipe & Structured Mesh & 2D & ESTIMATE VELOCITY & 1000 & 200\\
\bottomrule
\end{tabular}
}
\label{tab1}
\end{table}

First, we present two equations related to solid mechanics.

\textbf{Plastic Problem:} Consider a block of solid material \( \Omega = [0, L] \times [0, H] \) subjected to the impact of a frictionless rigid mould. The mould is geometrically parameterized by the function \( S_d \in H^1([0, L]; \mathbb{R}) \) and moves with a constant velocity \( v \). The lower edge of the material block is clamped, and a displacement boundary condition is imposed on the upper edge. The material behavior is described using an elastic-plastic constitutive model, where the plastic flow is governed by the relationship between the plastic multiplier and the stress tensor. The dataset for this problem comes from the baseline model geo-FNO, containing 980 training data points and 80 test data points. The data is distributed on a structured grid of size \( 101 \times 31 \), with 20 time steps.
 
\textbf{Elastic Problem:} The governing equation for solid mechanics is given by:
\begin{equation}
    \rho_s \frac{\partial^2 u}{\partial t^2} + \nabla \cdot \sigma = 0
\end{equation}
where \( \sigma \) is the stress tensor and \( \rho_s \) is the material density. Consider a unit plane \( \Omega = [0, 1] \times [0, 1] \) with an arbitrary-shaped void. The bottom edge is fixed, and a traction force \( t = [0, 100] \) is applied at the top. The material is described by the incompressible Rivlin-Saunders model. The data is generated using a finite element method with approximately 100 quadratic quadrilateral elements. The data is provided in point cloud format and is converted into a regular grid by differencing.
    
Below we give the darcy flow equation.

\textbf{Darcy Flow:} Henry Darcy's law describes the flow of fluids through porous media and is widely applicable in various fields such as heat conduction, electric current, and diffusion theory.

We consider the steady-state two-dimensional Darcy flow equation in a unit rectangle as a second-order, linear, elliptic partial differential equation with Dirichlet boundary conditions:

\begin{equation}
    -\nabla \cdot \left( a(x) \nabla u(x) \right) = f(x), \quad x \in (0,1)^2
\end{equation}
\begin{equation}
    u(x) = 0, \quad x \in \partial(0,1)^2
\end{equation}

Here, \(a\) is the diffusion coefficient and \(f\) is the forcing function. For this equation, our main focus is on obtaining the solution operator through the diffusion coefficient. The dataset used in this study comes from the baseline paper with a grid size of 421 × 421.

Below we give the Navier-Stokes equations and its two variant forms.  

\textbf{Navier-Stokes Equation(N-S):} The Navier-Stokes equation describes fluid motion. Consider the two-dimensional Navier-Stokes equation for a viscous, incompressible fluid on the unit ring surface, described in vorticity form as:

\begin{equation}
\frac{\partial \omega(x, t)}{\partial t} + u(x, t) \cdot \nabla \omega(x, t) = \nu \Delta \omega(x, t) + f(x), \quad x \in (0,1)^2, \ t \in (0, T]
\end{equation}

\begin{equation}
\nabla \cdot u(x, t) = 0, \quad x \in (0,1)^2, \ t \in (0, T]
\end{equation}

\begin{equation}
\omega(x, 0) = \omega_0(x), \quad x \in (0,1)^2
\end{equation}

Here, \( u \) is the velocity field, \( \omega \) is the vorticity, \( \omega_0 \) is the initial vorticity, \( \nu \) is the viscosity coefficient, and \( f \) is the external forcing term. The goal of the neural network is to learn the operator that maps vorticity at time \( t = 10 \) to vorticity at some later time \( T > 10 \).

The initial vorticity \( \omega_0(x) \) is generated by a specific random process, and the external forcing term is given by \( f(x) \). The equation is solved using a stream function representation combined with a pseudospectral method. Data is downsampled to a grid resolution of \( 64 \times 64 \) for training and testing.

\textbf{Airfoil Problem:} The Airfoil Problem is analyzed using Euler's equations for inviscid fluid flow, which are a simplified form of the Navier-Stokes equations. They describe high-speed flows where viscous effects are negligible. The equations are:

\begin{equation}
\frac{\partial \rho}{\partial t} + \nabla \cdot (\rho \mathbf{u}) = 0
\end{equation}

\begin{equation}
\frac{\partial (\rho \mathbf{u})}{\partial t} + \nabla \cdot (\rho \mathbf{u} \otimes \mathbf{u}) + \nabla p = 0
\end{equation}

\begin{equation}
\frac{\partial E}{\partial t} + \nabla \cdot (E\mathbf{u}) + p\nabla \cdot \mathbf{u} = 0
\end{equation}

Here, \(\rho\) is the mass density, \(\mathbf{u}\) is the velocity, \(p\) is the pressure, and \(E\) is the total energy. The far-field boundary conditions are \(\rho_1 = 1\), \(p_1 = 1.0\), \(M_1 = 0.8\), and \(\text{AoA} = 0\), with a no-penetration condition at the airfoil. The dataset, consisting of 1000 training and 200 test points, was generated using a second-order implicit finite volume solver with a C-mesh of 200 quadrilateral elements. The neural network learns the mapping from airfoil shape parameters to the surrounding flow field. 

\textbf{Pipe Problem:} The Pipe Problem involves studying fluid flow through a pipe governed by the Navier-Stokes equations. For steady, incompressible, and laminar flow, the equations simplify as:

\begin{equation}
\nabla \cdot \mathbf{v} = 0
\end{equation}

\begin{equation}
\frac{\partial \mathbf{v}}{\partial t} + (\mathbf{v} \cdot \nabla) \mathbf{v} = -\nabla p + \nu \nabla^2 \mathbf{v}
\end{equation}

Here, \(\mathbf{v}\) is the velocity, \(p\) is the pressure, and \(\nu = 0.005\) is the viscosity. At the inlet, a parabolic velocity profile with maximum velocity \(\mathbf{v} = [1; 0]\) is established. The outlet uses a free boundary condition, while the pipe surface has a no-slip boundary condition. The pipe has a length of 10 and a diameter of 1, with its centerline defined using piecewise cubic polynomials. The experiment uses 1000 training mesh points and 200 test mesh points. Neural networks are concerned with modeling the velocity field through geometric parameters, using grid point locations (129×129) and horizontal velocity as input and output data.

It is specifically noted that the pipe, airfoil, Elasticity-G, and Plasticity datasets were provided by Li et al. \citep{li2023fourier}, and the darcy and Navier-Stokes datasets were also provided by Li et al \citep{li2020fourier}.

\subsection{Results and discussion}

\subsubsection{Main results}

In this section, we focus on presenting the performance of the model across multiple experimental datasets. We first present a comparison of the experimental results under different equations, followed by a detailed discussion.

As shown in Table \ref{tab2}, DPNO achieved 5 optimal and 1 second-best results out of 8 methods, demonstrating outstanding performance. Compared to the suboptimal model, the LSM error is reduced by an average of 11.2\%, and the error is reduced by as much as 36.3\% compared to our backbone architecture, FNO.

\begin{table}[h!]
\centering
\caption{The MSE errors of different methods on various datasets, with the best results in bold and the second-best results underlined.}
\resizebox{\linewidth}{!}{%
\begin{tabular}{l c c c c c c c} 
\hline
Model & Pipe & Airfoil & Elasticity-G & Plasticity & Darcy & NS \\  
\hline
U-Net   & 0.0065 & 0.0079 & 0.0531 & 0.0051 & 0.0080 & 0.1982 \\
FNO & 0.0067 & 0.0138 & 0.0508 & 0.0074 & 0.0078 & 0.1556 \\
WMT     & 0.0077 & 0.0075 & 0.0520 & 0.0076 & 0.0082 & 0.1541 \\
U-NO    & 0.0100 & 0.0078 & 0.0469 & 0.0034 & 0.0183 & 0.1731 \\
U-FNO  & 0.0056 & 0.0269 & 0.0480 & 0.0039 & 0.0183 & 0.2231 \\
SNO & 0.0294 & 0.0893 &  0.0987 &  0.0070 & 0.2568 & 0.0495 \\
F-FNO   & 0.0070 & 0.0078 & 0.0475 & 0.0047 & 0.0077 & 0.2322 \\
LSM     & \textbf{0.0050} & \underline{0.0059} & \underline{0.0408} & \underline{0.0025} & \underline{0.0065} & \underline{0.1535} \\
DPNO          & \underline{0.0052} & \textbf{0.0054}(8.5\%) & \textbf{0.0375}(8\%) & \textbf{0.0020}(20\%) & \textbf{0.0057}(12.3\%) & \textbf{0.1423}(7\%) \\
\hline
\end{tabular}%
}
\label{tab2}
\end{table}

\begin{figure}[t] 
\centering 
\resizebox{\textwidth}{!}{
  \includegraphics{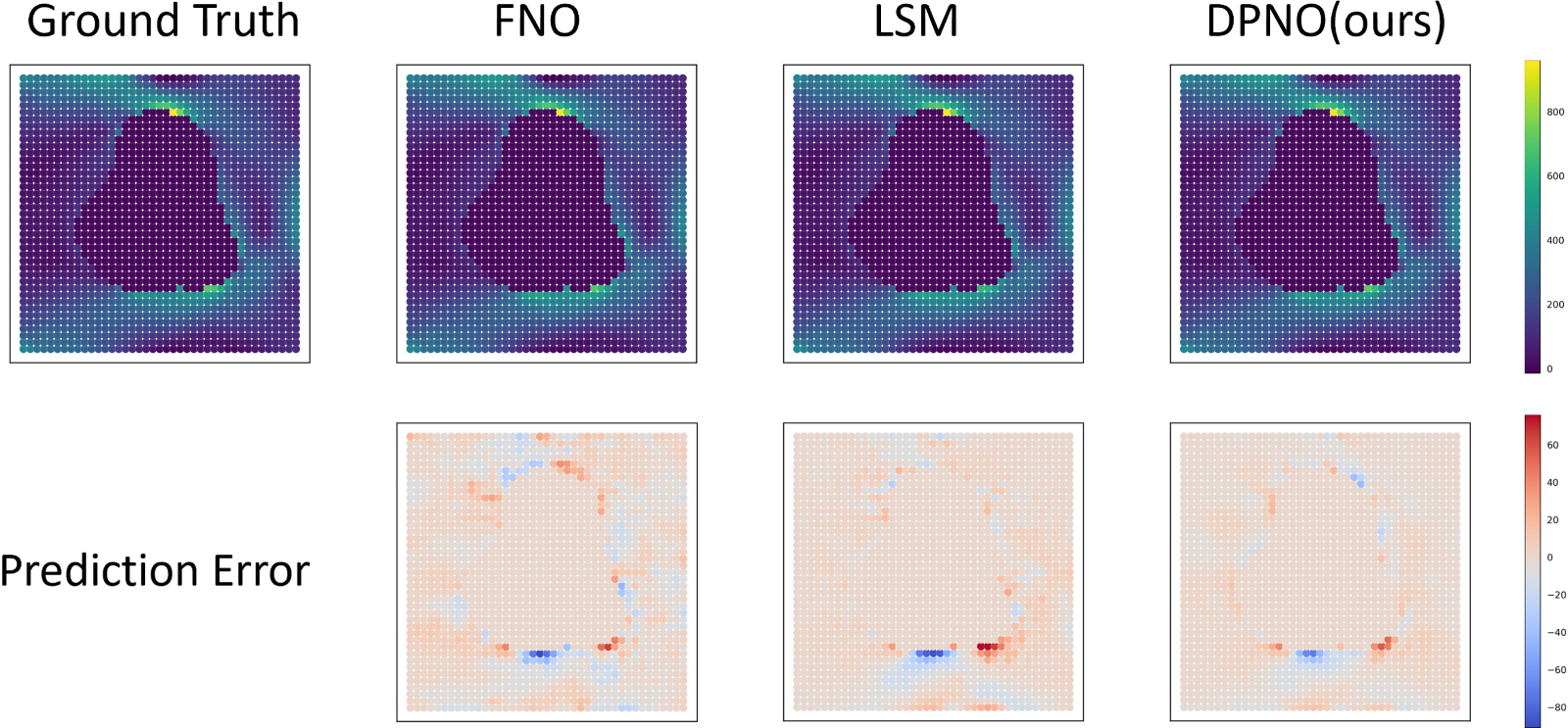} 
}
\caption{The first row shows the true values (with color representing the magnitude of stress) and the predicted values, while the second row shows the model's absolute error.}
\label{elasticity} 
\end{figure}

For the elasticity equation, the model's task is to learn the relationship between the material structure and stress, in order to estimate the stress distribution in the material or structure when subjected to an external force. For the plasticity equation, the model's task is to learn the mapping between the boundary conditions and grid displacements, in order to sequentially simulate the material deformation process. The visualization of the corresponding experimental results is shown in Figure \ref{elasticity}. Based on the figure, it can be seen that DPNO is able to effectively construct the two solid equation tasks.

\begin{figure}[t] 
\centering 
\resizebox{\textwidth}{!}{
  \includegraphics{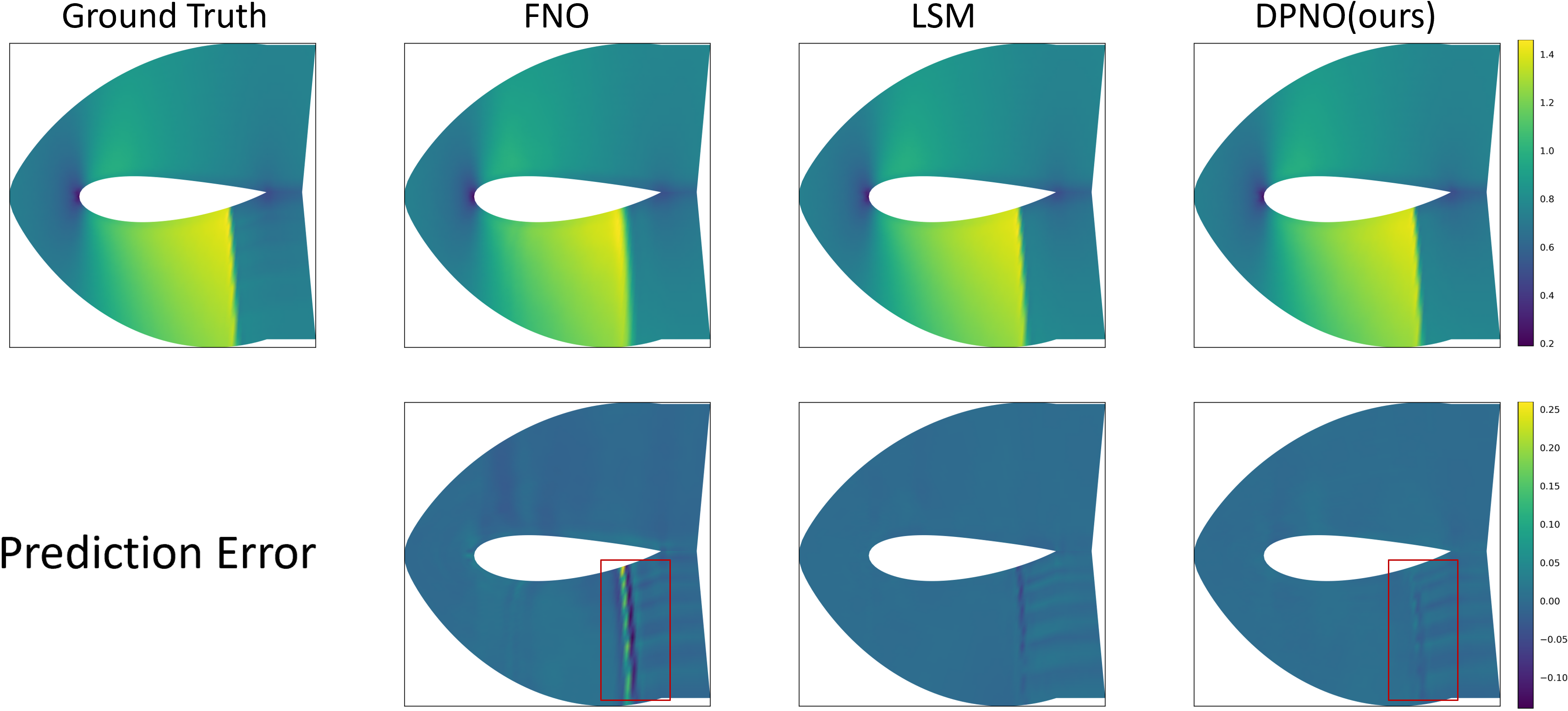} 
}
\caption{Airfoil: The first row shows the true values (with color representing the magnitude of the velocity.) and the predicted values, while the second row shows the model's absolute error.}
\label{airfoil} 
\end{figure}

\begin{figure}[t] 
\centering 
\resizebox{\textwidth}{!}{
  \includegraphics{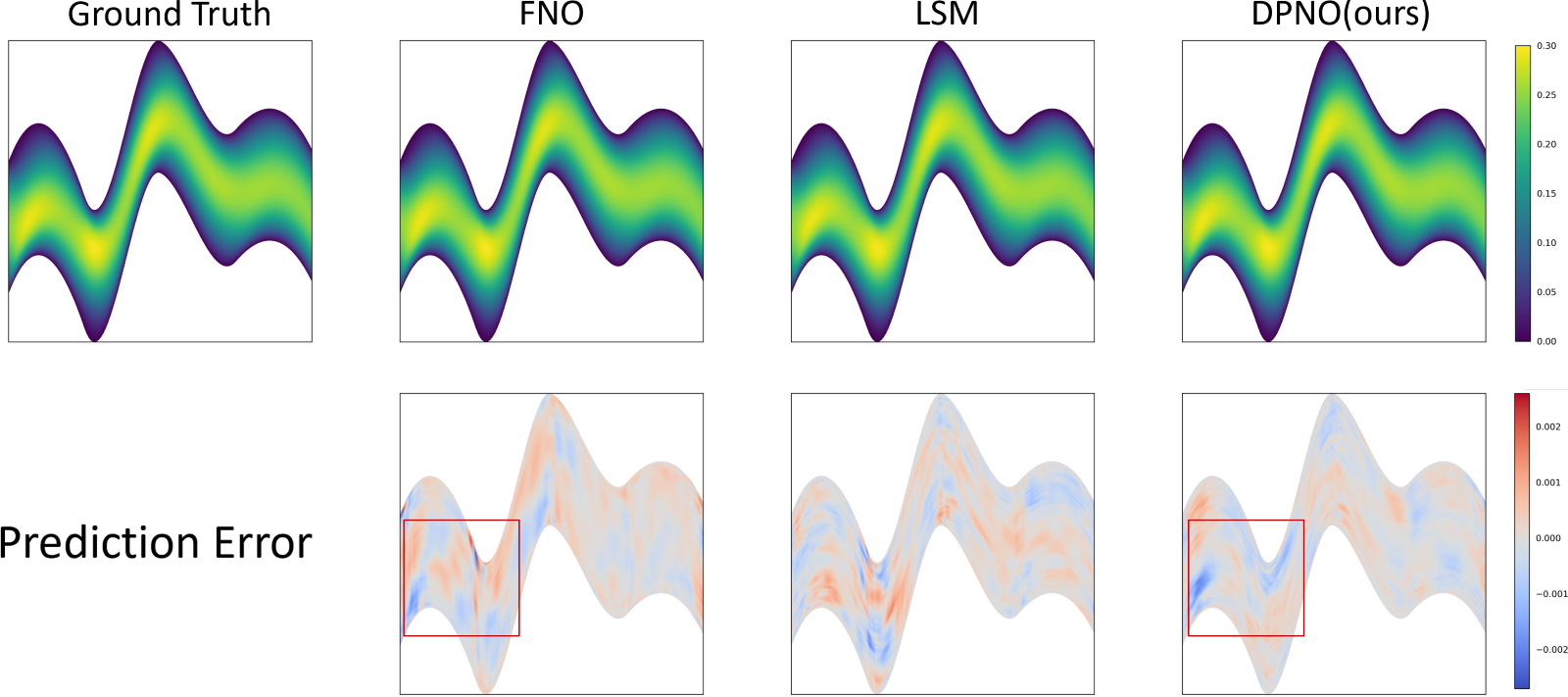} 
}
\caption{Pipe: The first row shows the true values (with color representing the magnitude of the velocity.) and the predicted values, while the second row shows the model's absolute error.}
\label{pipe} 
\end{figure}

For the airfoil and pipe problems, the model's main task is to learn the relationship between the material structure and the velocity field in order to construct the velocity field. As shown in Figure \ref{airfoil}, \ref{pipe}, the experimental results for the airfoil and pipe problems indicate that DPNO can effectively fit the regions of the velocity field with rapid changes (high-frequency information). This is primarily due to the design of our decoder architecture, which smooths high-frequency information, allowing the Fourier modes to capture this high-frequency data. In contrast, FNO exhibits significant fitting errors in this region, mainly due to information loss caused by mode truncation, as it has not learned this aspect well.

\begin{figure}[t] 
\centering 
\resizebox{\textwidth}{!}{
  \includegraphics{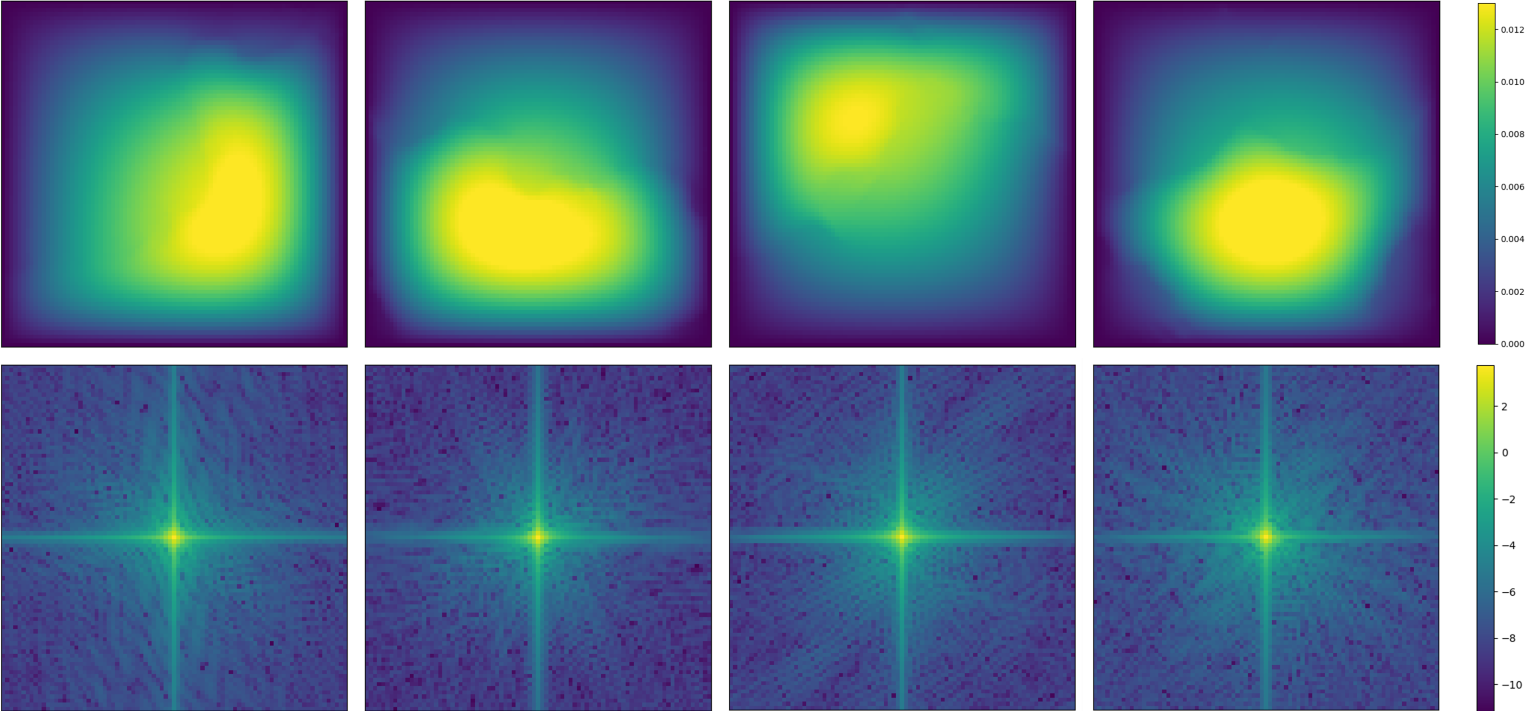} 
}
\caption{Some real values of the Darcy data and their frequency distributions, with the low frequencies being closer to the center point.}
\label{fig2} 
\end{figure}

Taking the Darcy dataset as an example, we will demonstrate the superiority of DPNO over FNO. First, we present several real images of different Darcy data along with their frequency distribution maps. These frequency distribution maps are extracted using the Fourier transform and have undergone a logarithmic transformation for better visualization, as shown in Figure \ref{fig2}. From the figures, it is evident that the majority of the information in the data consists of low-frequency components, which aligns with the statement made in the introduction. Therefore, enhancing the model's ability to learn low-frequency components is crucial for reducing errors.

\begin{figure}[t] 
\centering 
\resizebox{\textwidth}{!}{
  \includegraphics{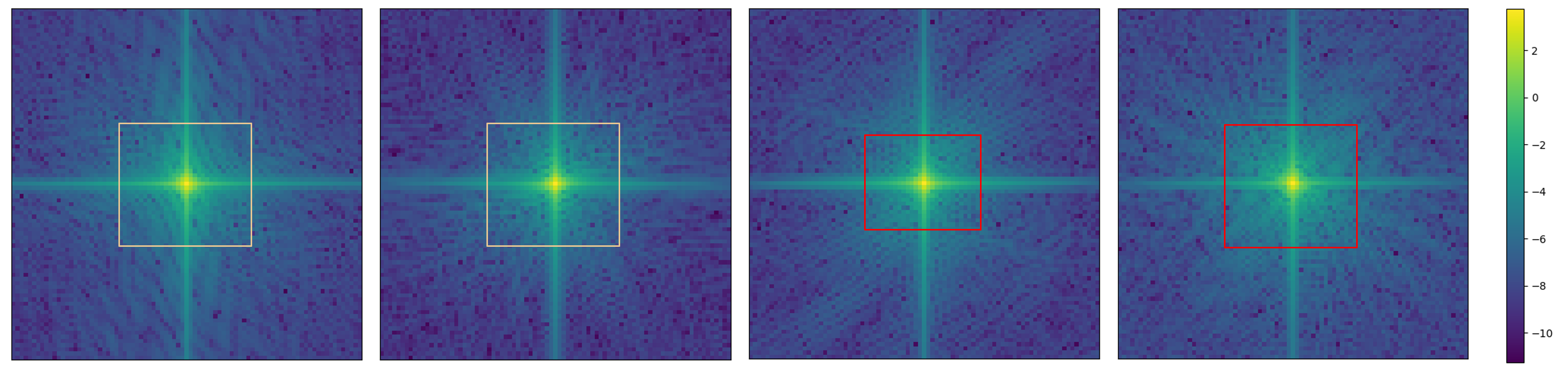} 
}
\caption{The two subplots on the left represent the range of learning frequency patterns of the two sub-Fourier layers in FNO, and the two subplots on the right represent the range of learning frequency patterns of the two sub-Fourier layers in DPNO.}
\label{fig3} 
\end{figure}

FNO is a serial iterative architecture that performs truncation of frequency patterns after Fourier transformation to learn low-frequency components. However, because different Fourier layers use the same set of frequency patterns, this leads to limited learning capability. In DPNO, we use downsampling to process frequency information at multiple scales, and within the same data, we employ parallel Fourier layers for learning. The two parallel Fourier layers use different frequency patterns, which better capture low-frequency components. To illustrate this more intuitively, we compare the use of two layers of Fourier layers, as shown in Figure \ref{fig3}. It can be seen that the range of frequencies learned by FNO is fixed, whereas the frequency range learned by DPNO not only includes the range learned by FNO but also enhances the learning of core low-frequency components.

\subsubsection{Experimental results at different resolutions}
In this section, we mainly demonstrate the model's adaptability at different resolutions. As the number of grids increases or decreases, the model performance remains effective. The given data consists of a 421x421 grid, and four different resolutions, \( s = 85, 141, 211, \) and 421, were obtained through downsampling. The experimental results are shown in Table \ref{tab3}.

\begin{table}[h]
\centering
\caption{Model performance comparison on the Darcy Flow benchmark under different resolutions.}
\begin{tabular}{l c c c} 
\hline
Model & 141*141 & 211*211 & 421*421 \\
\hline
FNO   & 0.0076 & 0.0073 & 0.0074\\
F-FNO  & 0.0063 & 0.0064 & 0.0065\\
U-FNO  & 0.0063 & 0.0073 & 0.0077\\
LSM  & 0.0063 & 0.0063 & 0.0068\\
DPNO  & 0.0059 & 0.0058 & 0.0062\\

\hline
\end{tabular}
\label{tab3}
\end{table}

\begin{figure}[t] 
\centering 
\resizebox{\textwidth}{!}{
  \includegraphics{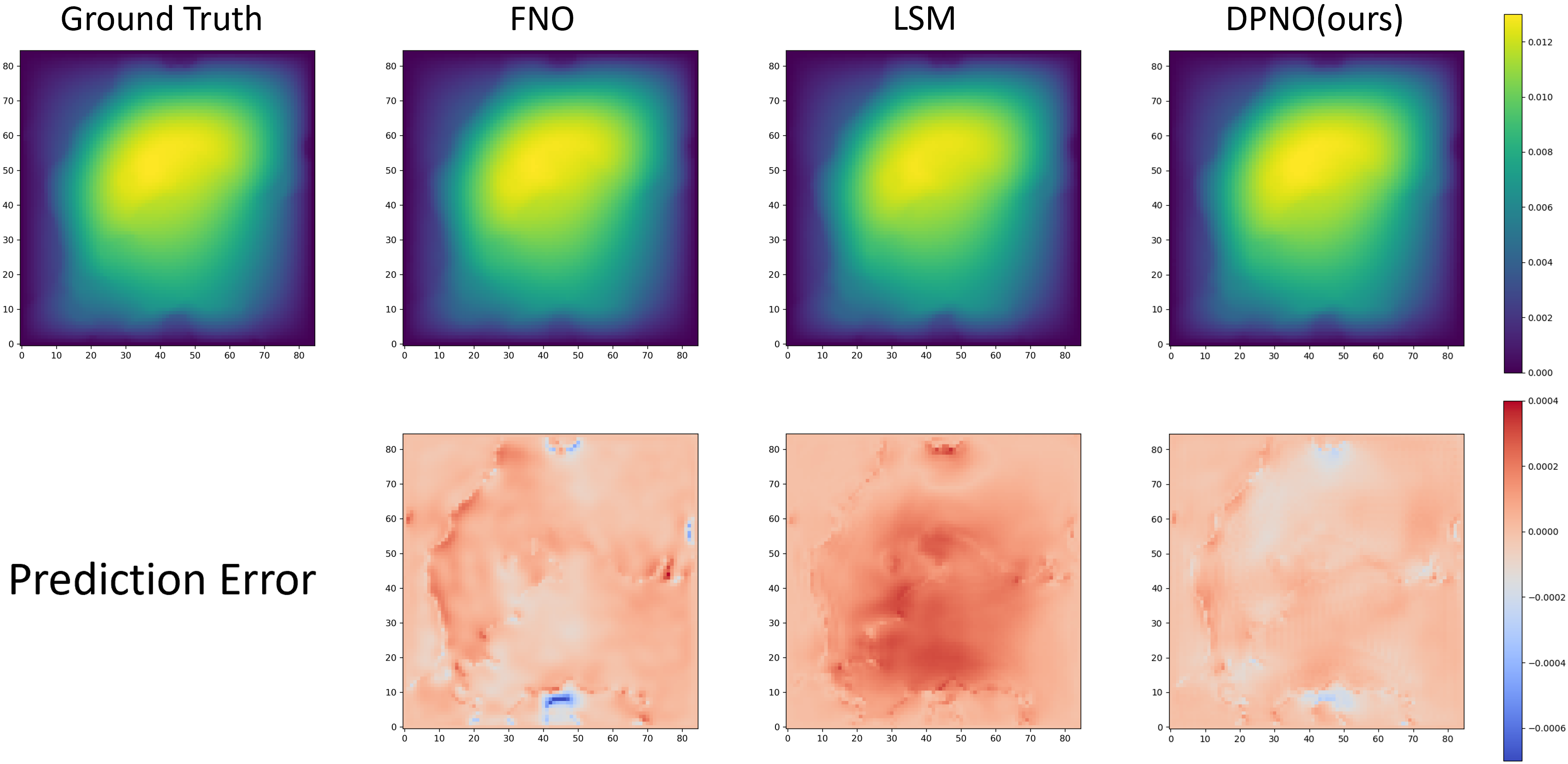} 
}
\caption{The distribution of the velocity field on an 85x85 grid in the Darcy dataset is shown. The first row presents the true values and the predictions from the main models, while the second row depicts the errors between the predicted values and the true values for each model.}
\label{fig4} 
\end{figure}

At different resolutions, we observe that DPNO performs the best among all baseline models, achieving the lowest relative error rates. Specifically, DPNO clearly outperforms FNO and its variants (including F-FNO and U-FNO), primarily because the latter use fixed frequency patterns. In contrast, DPNO learns and processes frequency information at multiple scales, using a hybrid approach of high and low frequency patterns, which allows it to better capture low-frequency components. DPNO demonstrates a stronger ability to learn low-frequency information, thereby improving the accuracy of the solution. To provide a more intuitive demonstration, we present the prediction and error maps of the model on some Darcy data, as shown in Figure \ref{fig4}.

\subsubsection{Zero-shot super-resolution}

Neural operators are invariant to mesh changes, meaning they can be trained on lower-resolution data and then evaluated at higher resolutions, even without prior exposure to higher-resolution data. Using the Darcy dataset as an example, we trained the model at a fixed resolution and tested it on data at other resolutions. The experimental results are presented in Table \ref{grid-resolutions}. It can be observed that DPNO performs well across various cross-experiments, achieving super-resolution in the spatial domain.

\begin{table}[ht]
\centering
\caption{The model is trained at a specified resolution, and the MSE error comparison is performed at multiple unseen resolutions during testing.}
\label{grid-resolutions}
\resizebox{\textwidth}{!}{
\begin{tabular}{|c|c|c|c|c|c|}
\hline
\multirow{2}{*}{Training grid resolution} & \multirow{2}{*}{\textbf{Model}} & \multicolumn{4}{c|}{\textbf{Test grid resolution}} \\ \cline{3-6}
                  &                                   & \textbf{85*85} & \textbf{141*141} & \textbf{211*211} & \textbf{421*421} \\ \hline

\multirow{2}{*}{85*85}   & FNO  & 0.0078 & 0.0365 & 0.0533 & 0.0711 \\ \cline{2-6}
                          & DPNO & 0.0057 & 0.0345 & 0.0535 & 0.0752 \\ \hline

\multirow{2}{*}{141*141} & FNO  & 0.0374 & 0.0071 & 0.0220 & 0.0413 \\ \cline{2-6}
                          & DPNO & 0.0379 & 0.0059 & 0.0216 & 0.0438 \\ \hline

\multirow{2}{*}{211*211} & FNO  & 0.0622 & 0.0245 & 0.0073 & 0.0251 \\ \cline{2-6}
                          & DPNO & 0.0585 & 0.0213 & 0.0058 & 0.0226 \\ \hline

\multirow{2}{*}{421*421} & FNO  & 0.0907 & 0.0514 & 0.0276 & 0.0074 \\ \cline{2-6}
                          & DPNO & 0.0719 & 0.0575 & 0.0315 & 0.0062 \\ \hline
\end{tabular}
}
\end{table}

\begin{figure}[t] 
\centering 
\resizebox{\textwidth}{!}{
  \includegraphics{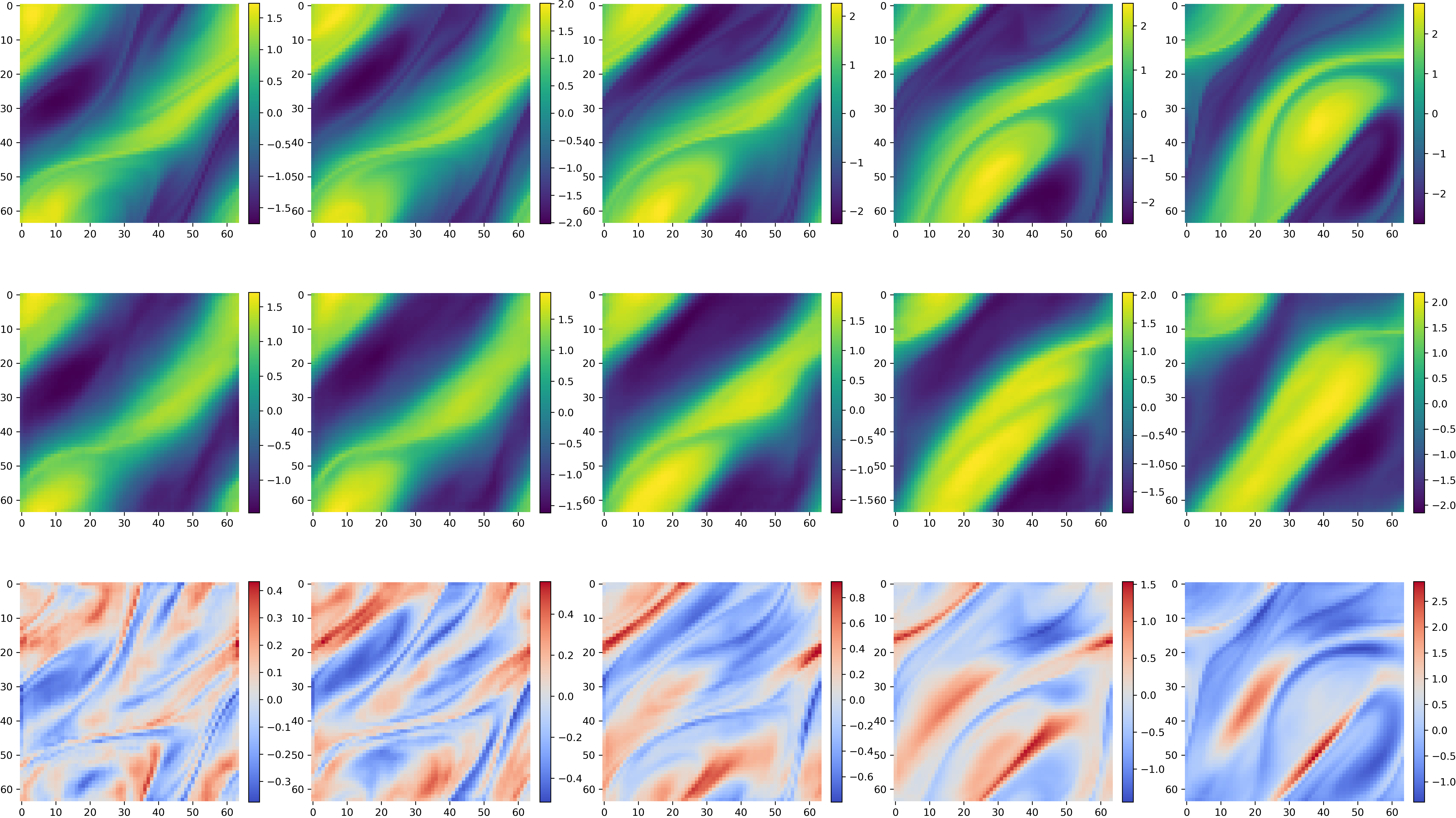} 
}
\caption{The results of the DPNO analysis are presented for the Navier-Stokes equation. The first line of the equation represents the true value(t=11,13,15,17,19), which is shown as the vorticity field distribution on a \( 64 \times 64 \) grid. The second line shows the model's prediction, and the third line indicates the absolute error. Each column corresponds to a specific point in time.}
\label{ns} 
\end{figure}

For the Navier-Stokes equation, the model focuses on learning the vorticity field transformation from time steps \text{0-T}, in order to predict the vorticity field change for \( t > T \). In this experiment, we use time steps 0 to 10 to train the model and predict the vorticity field change for time steps \text{11-20}. We provide a visualization of the DPNO experimental results, as shown in Figure \ref{ns}.

As demonstrated in Figure \ref{ns}, the DPNO model exhibits a high degree of efficacy in representing vorticity field variations. It is evident that the DPNO model has a commendable capacity for extrapolation. It should be noted that the DPNO model is constrained to learning alterations within the first 10 time steps; however, it demonstrates a remarkable ability to predict changes in the vorticity field beyond this time frame. This property signifies its super-resolution capability in the temporal domain.

In summary, the reason why DPNO is able to generalize well beyond the training range is primarily due to spectral transformations. Our method captures relevant physical information from a global perspective in the frequency domain by mapping data from the spatial or temporal domain to the frequency domain, which allows it to effectively learn the underlying physical laws.

\subsubsection{Ablation experiment}

Although the parallel operator block is not a traditional module in the strictest sense, we still compare it with the baseline model to demonstrate the effectiveness of this architecture. In the following discussion, we refer to a single Fourier transform and its inverse as a \textit{submodule}.

First, the parallel operator block learns the mapping relationship between input and output using two independent submodules, each of which learns relevant information at different frequency modes. In contrast, in FNO, the two submodules are interrelated and use fixed frequency modes for learning, resulting in a more limited set of frequency information being learned.

We perform an ablation experiment by transforming the parallel structure into a serial one, similar to FNO. The experimental results are shown in Table \ref{tab4}. As observed, the parallel architecture in DPNO architecture performs more effectively in learning the data.

\begin{table}[h]
\centering
\caption{Comparison of MSE for DPNO/P and DPNO on the 6 benchmark datasets. The improvement percentage is included in the DPNO row.}
\resizebox{\linewidth}{!}{%
\begin{tabular}{l c c c c c c} 
\hline
Model  & Pipe  & Airfoil & Elasticity-G & Plasticity & Darcy & NS  \\
\hline
DPNO/P & 0.0059 & 0.0061 & 0.0384 & 0.0021 & 0.0069 & 0.1586 \\
DPNO   & 0.0052 & 0.0054 & 0.0375 & 0.0020 & 0.0057 & 0.1423 \\
\hline
\end{tabular}
}
\label{tab4}
\end{table}

\subsubsection{Parameter count and time analysis}

In this section, we provide the parameter information of several main models as well as an analysis of their time performance. 

As shown in Table \ref{param-time}, we note that compared to FNO, the DPNO model has a larger number of parameters, primarily due to the design of the multi-scale architecture. However, while improving accuracy, it sacrifices some parameters, which is acceptable. At the same time, compared to LSM, DPNO has a similar number of parameters but exhibits better performance. In terms of training and testing time efficiency, since the data between different scales is parallelized in the frequency domain, DPNO achieves good training and testing speed. Overall, DPNO strikes a good balance between efficiency and accuracy, further proving the superiority of this method.

\begin{table}[h]
\centering
\caption{}
\resizebox{\textwidth}{!}{
\begin{tabular}{|c|c|c|c|c|c|c|c|c|c|c|c|c|}
\hline
\textbf{Data} & \multicolumn{3}{c|}{\textbf{Darcy}} & \multicolumn{3}{c|}{\textbf{NS}} & \multicolumn{3}{c|}{\textbf{Pipe}} & \multicolumn{3}{c|}{\textbf{Airfoil}} \\ \hline
\textbf{Model} & \textbf{Params} & \textbf{Train Time} & \textbf{Test Time} & \textbf{Params} & \textbf{Train Time} & \textbf{Test Time} & \textbf{Params} & \textbf{Train Time} & \textbf{Test Time} & \textbf{Params} & \textbf{Train Time} & \textbf{Test Time} \\ \hline
DPNO (ours)    & 10.5MB          & 2.8s                & 0.78s              & 8.4MB           & 12.9s               & 1.29s             & 13.3MB          & 6.5s               & 0.88s            & 14.2MB          & 4.1s               & 0.77s            \\ \hline
FNO            & 2.3MB           & 1.2s                & 0.65s              & 2.3MB           & 4.2s                & 0.91s             & 2.3MB           & 1.7s               & 0.75s            & 2.3MB           & 1.4s               & 0.67s            \\ \hline
LSM            & 18.3MB          & 5.4s                & 1s                 & 18.3MB          & 33.2s               & 2.56s             & 10.3MB          & 7.3s               & 1s               & 4.6MB           & 3.9s               & 0.79s            \\ \hline
\end{tabular}
}
\label{param-time}
\end{table}

\begin{figure}[t] 
\centering 
\resizebox{\textwidth}{!}{
  \includegraphics{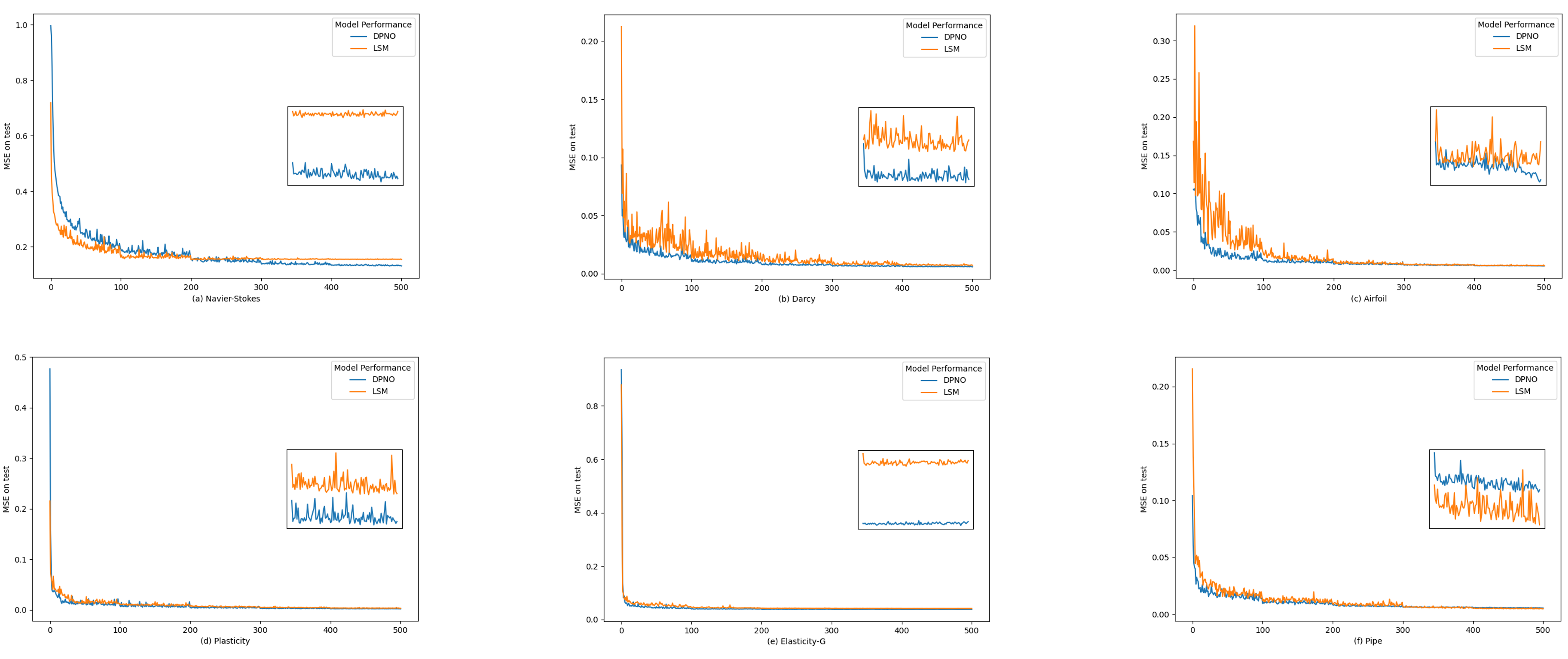} 
}
\caption{The loss curve of DPNO and LSM after 500 training epochs.}
\label{loss} 
\end{figure}

Finally, we present the loss curve during the experimental training process, as shown in Figure \ref{loss}. From the figure, it can be seen that DPNO is very stable during training and converges quickly.

\section{Conclusions}
Neural operators excel at learning functional spaces. In this paper, we propose a deeply parallel Fourier neural operator to address the issue of error accumulation during FNO iterations. The data are projected into different latent spaces through the decoder and learned in an end-to-end manner via multiple parallel operator blocks. Excellent performance is achieved on six benchmark datasets for partial differential equations, with an average improvement of 33\% compared to the baseline model, FNO.

Although deep learning has made significant progress in learning partial differential equations, there are still limitations in long-term prediction. Specifically, the time dependence is not captured effectively. As mentioned in Section 4.3.3 regarding the experiments on the Navier-Stokes equation, although DPNO has achieved good experimental results, the error continues to increase as the time steps increase. For dynamically changing systems, deep learning models may require constant updates and retraining to adapt to new conditions. Therefore, future work will focus on enhancing the neural operator model by integrating physical information models. By incorporating physical information, the data dependency of the neural operator model is reduced.

\section{Acknowledgments}
This work was supported by National Key Research and Development Project of China (2021YFA1000103, 2021YFA1000102), National Natural Science Foundation of China (Grant Nos.62272479, 62372469, 62202498), Taishan Scholarship (tstp20240506), Shandong Provincial Natural Science Foundation(ZR2021QF023). I would like to extend my sincere thanks to my teachers and the team for their valuable suggestions and experimental ideas in this research.








\end{document}